\documentclass[10pt, a4paper]{article}

\usepackage{lrec-coling2024} 
\usepackage{comment}
\usepackage{multirow}
\usepackage[acronym,nomain]{glossaries}
\newacronym{nlp}{NLP}{natural language processing}
\newacronym{pso}{PSO}{particle swarm optimization}
\newacronym{bert}{BERT}{Bidirectional Encoder Representations from Transformer}
\newacronym{sst2}{SST-2}{Stanford Sentiment Treebank2}
\newacronym{onnx}{ONNX}{Open Neural Network Exchange}
\newacronym{cv}{CV}{computer vision}
\usepackage{soul}
\usepackage{cleveref}

\title{ \textbf{The Impact of Quantization on the Robustness of Transformer-based Text Classifiers}}

\name{\large\bf Seyed Parsa Neshaei\textsuperscript{\textnormal{1}}, \large\bf Yasaman Boreshban\textsuperscript{\textnormal{1}}, \large\bf Gholamreza Ghassem-Sani\textsuperscript{\textnormal{1}},\\
and \large\bf Seyed Abolghasem Mirroshandel\textsuperscript{\textnormal{2}}} 

\address{\textsuperscript{\textnormal{1}}Computer Engineering Department, Sharif University of Technology, Tehran, Iran\\\textsuperscript{\textnormal{2}}Computer Engineering Department, University of Guilan, Rasht, Iran  \\
         \{seyed.neshaei, yasaman.boreshban, sani\}@sharif.edu\\ mirroshandel@guilan.ac.ir}

\abstract{
Transformer-based models have made remarkable advancements in various NLP areas. Nevertheless, these models often exhibit vulnerabilities when confronted with adversarial attacks. In this paper, we explore the effect of quantization on the robustness of Transformer-based models. Quantization usually involves mapping a high-precision real number to a lower-precision value, aiming at reducing the size of the model at hand. To the best of our knowledge, this work is the first application of quantization on the robustness of NLP models. In our experiments, we evaluate the impact of quantization on BERT and DistilBERT models in text classification using SST-2, Emotion, and MR datasets. We also evaluate the performance of these models against TextFooler, PWWS, and PSO adversarial attacks. Our findings show that quantization significantly improves (by an average of 18.68\%) the adversarial accuracy of the models. Furthermore, we compare the effect of quantization versus that of the adversarial training approach on robustness. Our experiments indicate that quantization increases the robustness of the model by 18.80\% on average compared to adversarial training without imposing any extra computational overhead during training. Therefore, our results highlight the effectiveness of quantization in improving the robustness of NLP models. 
\newline  }

\begin{document}
\maketitleabstract
\section{Introduction}
\label{sec:Introduction}

Pre-trained language models such as \gls*{bert}, T5, and Switch are vastly successful in various fields of \gls*{nlp}  \cite{devlin-etal-2019-bert,raffel2020exploring, wolf2020transformers,fedus2021switch}. However, these models still suffer from low robustness and high computational complexity \cite{jin-etal-2020-textfooler, cheng2020survey}.

Many researchers showed that these models are not robust enough, which means they are highly vulnerable to adversary examples. For this reason, the target of many studies is set to improve the robustness of neural network models \cite{gil-etal-2019-white, ren-etal-2019-generating}. One of the most successful techniques to overcome this problem is \textit{adversarial training}, which uses a set of generated adversarial attack examples for training the model to improve its robustness \cite{miyato2016adversarial, zhou2021defense}. One of the disadvantages of adversarial training is that it also increases the complexity of the training process
, and usually retains the vulnerability of the model against other types of attacks\cite{jia-liang-2017-adversarial, wang-bansal-2018-robust} 

Another concern about deep neural networks is that these models generally have millions of parameters \cite{devlin-etal-2019-bert,raffel2020exploring}, leading to a considerable inference time \cite{kim2021bert}. For this reason, these models are usually unsuitable for real-world and low-resource applications \cite{cheng2020survey}. Accordingly, various model compression methods have been introduced to address this issue \cite{oguntola2018slimnets}. One of these methods, which was also recently studied in the \gls*{cv} field \cite{galloway2017attacking, panda2019discretization, sen2020empir}, is called quantization. Quantized neural networks normally require fewer bits to represent the data structures such as weights and activation functions of the network. 
While previous studies showed reduced inference time, memory footprint, and low difference in the accuracy of quantized \gls*{nlp} models compared to their original counterparts \cite{kim2021bert}, the impact of quantization on the robustness of \gls*{nlp} model against adversarial examples has not yet been studied.

In this paper, we investigate the effect of quantization on the robustness of Transformer-based \gls*{nlp} text classifiers. For this purpose, we quantize two pre-trained models, \gls*{bert} and DistilBERT, and compare the accuracy of the original and quantized models. Then, to evaluate the effect of quantization on the robustness of the models, we employ Three different strong attack algorithms called TextFooler \cite{jin-etal-2020-textfooler}  \gls*{pso} \cite{zang-etal-2020-word} and PWWS \cite{ren-etal-2019-generating}. Finally, we compare the effect of the quantization approach with that of adversarial training. The results show that the robustness of the models on our generated adversarial examples is notably increased. In summary, the key contributions of this work are: 1) We propose quantization as a means of enhancing the \gls*{nlp} models against adversarial attacks. 2) We analyze the impact of quantization on various models and datasets against different attacks. 3) Although our experiments are performed on the task of text classification, we believe the proposed technique can be easily applied to other \gls*{nlp} tasks, too. 4) We demonstrate that in contrast with adversarial training, our approach significantly enhances the robustness of the model without imposing any extra overhead during training.

\section{Related Work}
\label{sec:Related-Work}

In the \gls*{nlp} applications, adversarial attacks are divided into three categories based on the perturbation levels: character level \cite{eger-etal-2019-text, he-etal-2021-model}, word level \cite{zang-etal-2020-word, alzantot-etal-2018-generating}, and sentence level \cite{wang-etal-2020-t3, huang-chang-2021-generating}. Currently, adversarial training is one of the most successful defense methods \cite{miyato2016adversarial, zhou2021defense}. 

 In recent years, some researchers in the area of \gls*{cv} showed that quantization improves the robustness of deep neural networks against adversarial attacks \cite{galloway2017attacking, panda2019discretization, chmiel2020robust, duncan2020relative}. It was also shown that using a combination of low-precision (i.e., quantized) and high-precision (i.e., original) models increases the robustness against adversarial data without reducing the accuracy on unperturbed data \cite{sen2020empir, pang2019improving}. Quantization was recently used in some \gls*{nlp} tasks as a model compression technique. It was claimed that this technique could significantly increase the efficiency of the model \cite{tao-etal-2022-compression,yang-etal-2022-compact,li-etal-2022-dq}. However, we still need to evaluate the effect of quantization on the robustness of \gls*{nlp} models.

\section{Datasets}
   We apply quantization to three datasets: \gls*{sst2} \cite{socher2013recursive}, Emotion \cite{saravia2018carer}, and MR \cite{Pang+Lee:05a}. 
   \gls*{sst2} and MR are binary sentiment analysis dataset that contains movie reviews. Emotion is also a sentiment analysis dataset based on Tweets. 
   
\section{Proposed Method}
\label{sec:Method}

Quantization is defined as a process of reducing the number of representation bits by means of using integers instead of floating point values. In Neural networks, which are vastly used in NLP applications, we normally employ 32 bits for representing different weights and activations by floating numbers. Both of these features can be quantized. Various quantization approaches have been explored in previous works. These approaches include linear quantization \cite{hubara2016binarized, hubara2017quantized, jacob2018quantization}, non-linear quantization \cite{li2019dimension}, and approximation-based methods \cite{lin2015neural}.
As an example, \cite{jacob2018quantization} have introduced a quantization scheme as an affine mapping of the form $$r = S(q-Z)$$ which maps each original value $r$ to its quantized equivalent $q$ with the \textit{quantization parameters} $S$ and $Z$. $S$ is known as the \textit{quantization scale} and $Z$ indicates the \textit{zero point} of the quantization process.

In this work, we use the \texttt{ONNXRuntime} Python library\footnote{https://onnxruntime.ai} to quantize two Transformer based models, by linearly mapping the floating point values of the models to an 8-bit quantization space. ONNXRuntime uses a similar equation to that of \cite{jacob2018quantization} for mapping the original values to the quantized ones. In the \textit{asymmetric quantization} process from \texttt{ONNXRuntime}, $S$ is calculated from the range of the original data as well as the quantization:
$$S = \frac{d_{max} - d_{min}}{q_{max} - q_{min}}$$
where $d_{max}$ and $d_{min}$ refer to the minimum and maximum of the data range, while $q_{max}$ and $q_{min}$ refer to the minimum and maximum of the quantization range.


In this paper, we aim to explore the effects of quantization on the outputs of Transformer-based models when faced with adversarial examples.
In our work, to find if quantization has any effect on how the models react to adversarial examples, we quantize two Transformer-based models, BERT \cite{devlin-etal-2019-bert} and DistilBERT \cite{sanh2019distilbert}, using the ONNX Python library. The details of the used models, along with the datasets on which the models were fine-tuned, are shown in Table \ref{tab:model-stats}. 
We use the \textit{dynamic} quantization method, which has been previously used \cite{zafrir2019q8bert} for quantizing Transformer-based models and calculates the values of $S$ and $Z$ on-the-fly.

\begin{table*}
\begin{tabular}{|l|c|c|l|c|c|}
\hline
\multirow{2}{*}{\textbf{Model Name}} &
        \multicolumn{2}{c}{\textbf{Model Binary Size (MB)}} &\multirow{2}{*}{\textbf{Dataset}} &
        \multicolumn{2}{c}{\textbf{Accuracy}} \\
&\textbf{Origin Model}         & \textbf{Quantized Model}       &  & \textbf{Origin Model} & \textbf{Quantized Model} \\ \hline
\multirow{3}{*}{BERT}    &  \multirow{3}{*}{417.90} & \multirow{3}{*}{173.42} & SST-2            & 94.12                                                                       & 93.52                                                                        \\ \cline{4-6} 
                            &                      &               & Emotion          & 92.60                                                                      & 91.90                                                         \\ \cline{4-6} 
                            &                  &                   & MR          & 86.49                                                                       & 86.21                                                                        \\ \hline     
\multirow{3}{*}{DistilBERT} & \multirow{3}{*}{255.54} & \multirow{3}{*}{132.46}  & SST-2            & 89.68                                                                       & 89.68                                                                        \\ \cline{4-6} 
                            &                  &                   & Emotion          & 92.70                                                                       & 92.55                                                            \\ \cline{4-6}        &                  &                   & MR          & 84.50                                                                       & 85.60                                                                        \\ \hline    
\end{tabular}
\caption{The Transformer-based models we used in our study. The values indicate minor losses in the accuracy of the quantized models, but a significant decrease in the model binary size.}
\label{tab:model-stats}
\end{table*}

To evaluate the effect of quantization on the robustness of the models, we aimed to generate adversarial examples for our models. To do so, we searched for attacks previously known to be effective in decreasing the accuracy of Transformer-based models, while also being relatively efficient regarding the time needed to generate the adversarial examples. We selected TextFooler \cite{jin-etal-2020-textfooler}, PWWS \cite{ren-etal-2019-generating}, and \gls*{pso} \cite{zang-etal-2020-word}, which claimed to be successful in tricking BERT models.

TextFooler, PWWS, and \gls*{pso} are  word-level attack algorithms. In these algorithms, words are initially sorted according to their impact on the output of the model. For this purpose, in turn, one word is removed from the sentence and the amount of change in the output of the model is calculated. Besides, the synonyms of each word are retrieved from the embedding space. Then, using these synonyms, a number of adversary sentences are created. The main difference between these algorithms is that TextFooler and PWWS select the final adversary sentences based on a greedy search, whereas \gls*{pso} employs an evolutionary algorithm for this task \cite{jin-etal-2020-textfooler,zang-etal-2020-word}. TextAttack \cite{morris-etal-2020-textattack} is a Python library incorporating various attack recipes including TextFooler, PWWS, and \gls*{pso} plus a set of \gls*{nlp} data augmentation tools. In this research, adversarial examples were generated using TextAttack.

\section{Experimental Results}

\subsection{Implementation Details}
\label{appendix:expsetup}
To evaluate if quantization affects the outputs of the models when faced with adversarial examples, we used four models fine-tuned in advance from HuggingFace\footnote{https://huggingface.co/}. For the BERT base models, we used \texttt{bert-base-uncased-sst2} for the SST-2, and \texttt{bert-base-uncased-emotion} for the Emotion dataset.
On the other hand, for the DistilBERT base models, we used \texttt{distilbert-base-uncased-sst2} and \texttt{distilbert-base-uncased-emotion} for the datasets of SST-2 and Emotion, respectively. 
All models are \textit{base} models, leading to 110 million parameters for the two BERT models \cite{devlin2018bert} and 66 million parameters for the two DistilBERT models \cite{sanh2019distilbert}. To make sure the models are fine-tuned correctly, we evaluated them using the \texttt{Evaluate} library by HuggingFace. The results, as shown in Table \ref{tab:model-stats}, indicate high accuracies and thus correct fine-tuning of the models we used.

To evaluate the effect of quantization, we generated the adversarial examples using TextAttack in the Google Colab\footnote{https://colab.research.google.com/} infrastructure. Generating adversarial examples took one hour on average using TextFooler and four hours on average using \gls*{pso}. After generating the examples, we used them as inputs to both the quantized and original models and measured their performance and accuracy on the perturbed dataset. We used each of the models given the adversarial examples for inference once and recorded the results.

\subsection{Results}

\begin{table*}
\centering
\begin{tabular}{|l|c|c|c|c|c|c|}
\hline

\multirow{2}{*}{\textbf{Model}} & \multicolumn{3}{c}{\textbf{SST-2}} & \multicolumn{3}{c}{\textbf{Emotion}}  \\
 & \textbf{TextFooler} & \textbf{PSO} & \textbf{PWWS}& \textbf{TextFooler} & \textbf{PSO}& \textbf{PWWS} \\
\hline
{BERT} & 5.99 & 6.96 & 12.00 & 2.00 & 7.78 & 6.90 \\
{BERT + Quantization} &  \bfseries 27.00 & \bfseries 16.04 & \bfseries 55.70 & \bfseries 19.90 &  \bfseries 13.53& \bfseries 29.30 \\
\hline
{DistilBERT} & 3.69 & 6.96 & 14.20 & 1.51 & 5.66 & 5.40 \\
{DistillBERT + Quantization} & \bfseries 26.95 & \bfseries 16.54 & \bfseries 45.50 & \bfseries 15.71 & \bfseries 9.74 & \bfseries 27.30 \\
\hline
\end{tabular}

\caption{After attack accuracy of BERT and DistillBERT models against the TextFooler, PSO, and PWWS attacks on the SST-2 and Emotion datasets.
}
\label{tab:afterattack-accuracy}
\end{table*}

Table~\ref{tab:model-stats} shows the effect of quantization on the size and accuracy of the model. The results show that by applying quantization to \gls*{bert} and DistilBERT, the size of each model is reduced to about 41.50\% and 51.80\% of the initial size of the model, respectively. Table~\ref{tab:afterattack-accuracy} shows the accuracy of these models after they are attacked by TextFooler, \gls*{pso}, and PWWS  methods to the \gls*{sst2} and Emotion datasets. The results show that quantization significantly increases the robustness of the model in all three types of attacks. As can be seen, despite the substantial reduction in the size and the low difference in the accuracy of the quantized and original models, the quantization of the models led to a considerable increase in their robustness.

Table~\ref{tab:afterattack-accuracy} shows that in the case of using the SST-2 dataset, the result of the TextFooler attack on \gls*{bert} and DistilBERT respectively shows an increase of 21.01\% (i.e., 27.00 minus 5.99) and 23.26\% (i.e., 26.95 minus 3.69) in terms of accuracy. Similarly, the result of the \gls*{pso} attack on these models respectively shows an increase of 9.08\% (i.e., 16.04 minus 6.96) and 9.58\% (i.e., 16.54 minus 6.96). Besides, the result of the PWWS attack on these models respectively shows an increase of 43.70\% (i.e., 55.70 minus 12.00) and 31.30\% (i.e., 45.50 minus 14.20). Likewise, while using the Emotion dataset, the result of  the TextFooler attack on these models respectively shows  an increase of 17.90\% (i.e., 19.90 minus 2.00) and 14.20\% (i.e., 15.71 minus 1.51). Similarly, the result of the \gls*{pso} attack on these models respectively shows an increase of 5.75\% (i.e., 13.53 minus 7.78) and 4.08\% (i.e., 9.74 minus 5.66). Finally, the result of the PWWS attack on these models respectively shows an increase of 22.40\% (i.e., 29.30 minus 6.90) and 21.90\% (i.e., 27.30 minus 5.40). As these experiments show, quantization has a significant positive impact on the robustness of the models against all types of attacks. In the majority of cases, the impact has been more substantial on improving the robustness of the \gls*{bert} model. That is perhaps because DistillBERT is already a compressed version of the \gls*{bert} model, which indicates quantization is more effective on larger models. In all the experimental results shown in Table~\ref{tab:model-stats} and Table~\ref{tab:afterattack-accuracy}, the evaluation  has been based on using one thousand test samples.

\begin{table}
\centering
\begin{tabular}{|l|c|c|}
\hline
\textbf{Model} & \textbf{Emotion} & \textbf{MR} \\
\hline
BERT  & 2.00 & 5.25  \\
BERT + Adv. Training & 5.25 & 18.70  \\
BERT + Quantization  & \bfseries 19.90 & \bfseries  41.65  \\
\hline

\hline
\end{tabular}

\caption{A comparison between after-attack accuracy of quantization and that of the adversarial training  method applied to the BERT model using the Emotion and MR dataset against the TextFooler attack. 
}
\label{tab:adversarial-training}
\end{table}
In Table~\ref{tab:adversarial-training}, we compare the effect of quantization on the robustness of the \gls*{bert} model with that of adversarial training. In this experiment, BERT is attacked by applying the TextFooler algorithm to the Emotion and MR datasets. For this purpose, Ten percent of the  training data is used to create adversary samples using TextFooler. These samples are then added to the training set. As Table~\ref{tab:adversarial-training} shows, in the case of using the Emotion dataset, the accuracy of the model is improved by 3.25\% (i.e., 5.25 minus 2.00); whereas by quantization, the accuracy is improved by 17.90\% (i.e., 19.90 minus 2.00). Similarly, in the case of using the MR dataset, the accuracy of the model is improved by 13.45\% (i.e., 18.70 minus 5.25); whereas by quantization, the accuracy is sharply improved by 36.40\% (i.e., 41.65 minus 5.25). Therefore, based on the results of our experiments, quantization is substantially more effective than adversarial training in improving robustness. Besides, the improvement by quantization is gained without imposing any extra computational overhead on the training process.
As the parameters of the models are represented with lower precision by the quantization process, this makes the quantized models less sensitive to the perturbation of the inputs. As a result, the quantized models achieve higher accuracy when dealing with adversarial data.



\subsection{Qualitative Examples}
\label{appendix:qexamples}

We present some adversarial examples generated by the attacks on selected sentences from the datasets:

\begin{itemize}
    \item The input \textit{"much of the way, though, this is a refreshingly novel ride"}, from the SST-2 dataset, is changed to \textit{"practically of the direction, though, this is a refreshfully novel twit"}. The BERT model is tricked into classifying the perturbed input as 0 (negative), but the quantized model correctly classifies it as 1 (positive).
    \item The input \textit{"i do feel insecure sometimes but who doesnt"}, from the Emotion dataset, is changed to \textit{"i do flavour unsafe sometimes but who doesnt"}. The DistilBERT model is tricked into classifying the perturbed input as "sadness", but the quantized model correctly classifies it as "fear".
\end{itemize}

\section{Conclusion}
In this paper, we evaluated the impact of quantization on the robustness of Transformer-based \gls*{nlp} models. We applied quantization to the BERT and DistilBERT models on the Emotion, \gls*{sst2}, and MR datasets against three adversarial attacks named TextFooler, \gls*{pso}, and PWWS. We showed that the accuracy of the BERT and DistilBERT models will only slightly decrease (by an average of 0.98\%) after the quantization. On the other hand, quantization significantly increased (by an average of 18.68\%) the average adversarial accuracy of the model. We also compared the effect of quantization on the robustness with that of adversarial training. Finally, we showed that, for the goal of improving the robustness of the model, quantization is significantly more effective than adversarial training. Besides, this advantage of quantization is obtained without imposing any extra computation overhead in training. In our future work, we intend to investigate the effect of quantization on newer pre-trained models such as T5 and Switch. We will also evaluate the impact of Ensembles of low-precision and high-precision models on the robustness.

\section*{Limitations}
Although we showed that the application of quantization in the text classification increases the after-attack accuracy, by no means, it is sufficient as a sole tool for providing robustness for a real-world application with an acceptable level of accuracy. Therefore, we must find ways of combining quantization with other methods to reach the accuracy of the original models on unperturbed data. Besides, to further evaluate the impact of quantization on the robustness, accuracy, and other features of NLP models or other NLP tasks, we need to perform more experiments using other datasets and attack algorithms. Our experimental goals were highly inflicted by the lack of the required hardware resources.

\section*{Ethics Statement}
Our work is pure research and it is not associated with any specific application. Therefore, we do not foresee any ethical concerns about the algorithms and resources used in the work. We have utilized publicly available datasets and open-source libraries, which have been published before.

Regarding the potential risks of misusing the results of our work, there is in general a possible risk of malicious usage of any research. However, the main goal of our research is to enhance the robustness of NLP models so that they can be less vulnerable to potential misuse.

\nocite{*}
\section{Bibliographical References}\label{sec:reference}

\bibliographystyle{lrec-coling2024-natbib}
\bibliography{lrec-coling2024}

\end{document}